\documentclass[letterpaper, 10 pt, conference]{ieeeconf}
\usepackage{times}
\usepackage[pdftex]{graphicx}
\usepackage{}
\usepackage{amsmath,amssymb,amsopn,amstext,amsfonts}
\usepackage{pifont}
\usepackage{cancel}
\usepackage{cite}
\usepackage{pdfsync}
\usepackage{balance}
\usepackage{color}
\usepackage{mathtools}
\usepackage{algpseudocode}
\usepackage{textcomp}

\usepackage[ruled,vlined,linesnumbered]{algorithm2e}

\algnewcommand\algorithmicforeach{\textbf{for each}}
\algdef{S}[FOR]{ForEach}[1]{\algorithmicforeach\ #1\ \algorithmicdo}

\usepackage{bm}

\usepackage{diagbox}
\usepackage{pifont}
\usepackage{amsmath}
\usepackage{multirow}
\usepackage{url}
\usepackage{verbatim}
\usepackage{footmisc}
\usepackage[linkcolor=blue,citecolor=blue,urlcolor=blue,colorlinks=true]{hyperref}

\usepackage{color}
\usepackage{tabularx}
\usepackage{float}
\usepackage{threeparttable}
\usepackage{booktabs}
\graphicspath{{figures/}}
\DeclareGraphicsExtensions{.png,.jpg,.eps}
\IEEEoverridecommandlockouts
\title{\LARGE \bf
    \emph{i-Octree}: A Fast, Lightweight,  and Dynamic  Octree for Proximity  Search
}
\author{
	Jun Zhu$^1$, Hongyi Li$^1$, Zhepeng Wang$^2$, Shengjie Wang$^{3,\dagger}$,    and Tao Zhang$^{1,\dagger}$,  \emph{Fellow, IEEE}% stops a space
	\thanks{$\dagger$ Corresponding author: {\tt\small taozhang@tsinghua.edu.cn}, {\tt\small wangsj23@mail.tsinghua.edu.cn}.}
	\thanks{$^1$ Department of Automation, Tsinghua University.  $^2$ Lenovo Research. $^3$ Institute for Interdisciplinary Information Sciences (IIIS), Tsinghua University.}
}%

\begin{document}
\maketitle
\newcommand{\note}[1]{\textcolor{red}{\emph{\bf#1}}}
\newcommand\footnoteref[1]{\protected@xdef\@thefnmark{\ref{#1}}\@footnotemark}
\newlength{\bibitemsep}\setlength{\bibitemsep}{.0238\baselineskip}
\newlength{\bibparskip}\setlength{\bibparskip}{0pt}
\let\oldthebibliography\thebibliography
\renewcommand\thebibliography[1]{%
\oldthebibliography{#1}%
\setlength{\parskip}{\bibitemsep}%
\setlength{\itemsep}{\bibparskip}%
}
\vspace{-1.0cm}
\begin{abstract}
Establishing the correspondences between newly acquired points and historically accumulated data (i.e., map) through nearest neighbors search is crucial in numerous robotic applications.
However, static tree data structures are inadequate to handle large and dynamically growing maps in real-time.
To address this issue, we present the \emph{i-Octree}, a dynamic octree data structure that supports both fast nearest neighbor search and real-time dynamic updates, such as point insertion, deletion, and on-tree down-sampling. 
The \emph{i-Octree} is built upon a leaf-based octree and has two key features: a local spatially continuous storing strategy that allows for fast access to points while minimizing memory usage, and local on-tree updates that significantly reduce computation time compared to existing static or dynamic tree structures.
The experiments show that \emph{i-Octree} outperforms contemporary state-of-the-art approaches by achieving, on average, a 19\% reduction in runtime on real-world open datasets.
%The experiments demonstrate that \emph{i-Octree} outperforms the state-of-the-art methods  by a over 30\% run-time reduction on  randomized data and  50\% on real-world open datasets.

%Our experiments demonstrate that \emph{i-Octree} significantly reduces runtime in various tasks, outperforming state-of-the-art methods on both randomized data and real-world open datasets.
\end{abstract}

\section{Introduction}\label{sect_intro}
% 核心还是NNS，以前的方法有什么问题
% 方法本身有什么需求，显存的解决方案有什么问题，我的解决方案又有什么优势
Nearest neighbors search (NNS) is necessary in many robotic applications, such as real-time LiDAR-based simultaneous localization and mapping (SLAM) and motion planning, where the data is sampled sequentially and real-time mapping is essential. 
For example, in LiDAR-based SLAM, NNS is crucial to compute feature\cite{Xiong_et_2011_, Behley_et_2012_histogram_descriptors}, estimate normal\cite{Mitra_Nguyen_2003, Klasing_et_2008}, and match new points to the map\cite{zhang2014loam, shan2018lego, lin2020loamlivox, xu2020fastlio}.
Recent advances in LiDAR technology have not only significantly reduced the cost, size, weight, and power of LiDAR  but also enabled high performance of LiDAR, making it an essential sensor for robots\cite{FAST-LIO2}.
However, this also poses challenges to NNS.
The current LiDAR sensor can produce a large amount of 3D points with centimeter accuracy at hundreds of meters measuring range per second.
The large amount of sequential data requires  being processed in real-time, which is a quite challenging task on robots with limited onboard computing resources.
To guarantee the efficiency of NNS, maintaining a large map supporting high efficient inquiry and dynamic updates with newly arriving points  in real-time comes to vital importance. 

Although various static tree data structures have been proposed, they struggle to meet the demands.
%These trees build spatial indices for data points through either partitioning the data or splitting the space. 
R-tree\cite{R_tree_Guttman_1984} and R$\ast$-tree\cite{R_star_tree_Beckmann_et_1990} partition the data by grouping nearby points into their minimum bounding rectangle in the next higher level of the tree. 
k-d tree\cite{bentley1975kdtree} is a well-known instance to split the space.
The octree  recursively splits the space equally into eight axis-aligned cubes, which form the volumes represented by eight child nodes.
Although k-d tree is a favored data structure in general k-nearest neighbors (KNN) search libraries,  it is hard to draw any final conclusions as to whether the k-d tree is better suited for NNS compared to other data structures.
Comparative studies\cite{vermeulen2017comparative, Elseberg_et_2012_comparision} show that the performance of different implementations of k-d trees can be diverse, and the octree is amongst the best performing algorithms especially for radius search due to its regular partitioning of the search space.
Despite its performance, octree has not been fully exploited.
%Hornung et al.\cite{Hornung_et_2013_octomap} proposed an octree based framework for the representation of 3D environments.
Elseberg et al. \cite{Elseberg_et_2013one} proposed an efficient octree to store and compress 3D data without loss of precision.
Behley et al. \cite{Behley_et_2015_octree} proposed an index-based octree that significantly improves the radius neighbor search in three-dimensional data, while the KNN search and dynamic updates are not enabled.
When incorporating these static trees in real robotic applications, re-building the entire tree from scratch\cite{rusu20113d} repeatedly is inevitable, which is so  time-consuming that the robots may fail to run.

In this paper, we propose a dynamic octree structure called \emph{i-Octree}, which incrementally updates the octree with new points and enables fast NNS.
In addition, our  \emph{i-Octree} boasts impressive efficiency in both time and memory, adaptable to various types of points, and allows for on-tree down-sampling and box-wise delete.
%Furthermore, the \emph{i-Octree} keeps the spatial characteristics of the 3D point data, making it easier to incorporate spatial features of local points.
%The state-of-the-art performance of \emph{i-Octree} is  verified on both random data and real-world point-cloud in LiDAR-based mapping applications. 
We conduct validation experiments on both randomized data and real-world open datasets to assess the effectiveness of \emph{i-Octree}. In the randomized data experiments, our \emph{i-Octree} demonstrates significant improvements in runtime compared to the state-of-the-art incremental k-d tree (i.e., \emph{ikd-Tree}\cite{FAST-LIO2} proposed recently). Specifically, it reduces run-time by 64\% for building the tree, 66\% for point insertion, 30\% for KNN search, and 56\% for radius neighbors search.
Moreover, when applied to real-world data in LiDAR-based SLAM, \emph{i-Octree} showcases remarkable time performance enhancements. It achieves over twice the speed of the original method while often maintaining higher accuracy levels.
Furthermore, Our implementation of  \emph{i-Octree} is open-sourced on Github\footnote{\url{https://sites.google.com/view/I-Octree}}.

The remaining paper is organized as follows: the design of \emph{i-Octree} is described in Section \ref{sec:implementation}.  Experiments are shown in Section \ref{sec:experiments}, followed by conclusions in Section \ref{sec:conclusion}.

\section{\emph{i-Octree} Design and Implementation}\label{sec:implementation}
\begin{figure*}[htb]
	\setlength\abovecaptionskip{-0.1\baselineskip}
	\centering
	\includegraphics[width=0.95\textwidth]{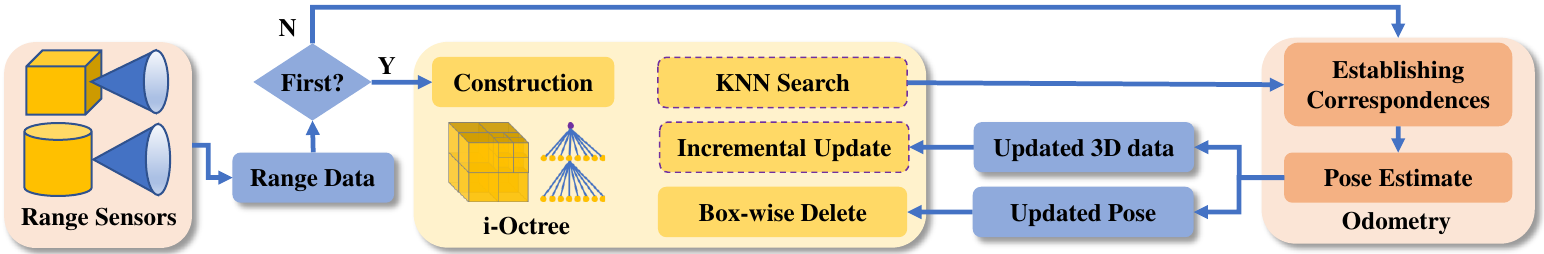}
	\caption{An example of using \emph{i-Octree} in odometry. 
		The \emph{i-Octree} and odometry collaborate to estimate the poses of the 3D data obtained from the range sensors. 
%		This collaboration enables the efficient and accurate estimation of poses, which is crucial for tasks such as SLAM and other 3D data-based applications. 
		The \emph{i-Octree} provides a robust and efficient data structure for storing and querying the 3D data, while the odometry  enables the estimation of the poses of the data points. 
%		Together, they form a powerful tool for 3D data processing and analysis.
	}
%	 做一些描述
	\label{fig:framework}    
	\vspace{-0.4cm}
\end{figure*}
The \emph{i-Octree} takes sequential point clouds as input with two objectives: dynamically maintains a global map and performs fast NNS (i.e., KNN search and radius neighbors search) on the map.
Fig. \ref{fig:framework} illustrates a typical application of the \emph{i-Octree}.
%The range sensors perceive their surrounding environment and generate sequential 3D range data periodically.
%Then the first scan of the range data will be used to construct the \emph{i-Octree} and define the global coordinate frame.
%The correspondences of the newly arriving data with the historical data are established by the \emph{i-Octree} through KNN search or radius neighbors search.
%Based on the correspondences, the poses of the new data can be estimated and the 3D points with pose will be added to the \emph{i-Octree}.
%To prevent the size of the map in \emph{i-Octree} from going unbound, only map points in a large local region (i.e., axis-aligned box) around the current position are maintained.
The range sensors continuously perceive their surroundings and generate sequential 3D range data periodically. 
The initial scan of the range data is utilized to construct the \emph{i-Octree} and define the global coordinate frame. 
The \emph{i-Octree} then facilitates the establishment of correspondences between the newly arriving data and the historical data through KNN search or radius neighbors search. 
Based on these correspondences, the poses of the new data can be estimated, and the 3D points with pose are added to the \emph{i-Octree}. 
To prevent the map size in \emph{i-Octree} from growing uncontrollably, only map points within a large local region (i.e., axis-aligned box) centered around the current position are maintained.
In the following, we first describe the data structure and construction of the \emph{i-Octree}, then we focus on dynamic updates and NNS.

\subsection{Data Structure and Construction}\label{subsec:construction}
%(变量较多的话可以附个表格介绍， 算法写的比较多 是都有创新吗？ 如果只有部分的话 建议就只写有创新的部分)
% 加点介绍
An \emph{i-Octree} node has up to eight child nodes,  each corresponding to one octant of the overlying axis-aligned cube.
An octant, starting with an axis-aligned bounding box with center $\boldsymbol{c} \in \mathbb{R}^3$ and equal extents $\boldsymbol{e} \in \mathbb{R}$, would be subdivided recursively into  smaller octants of extent $\frac{1}{2}\boldsymbol{e}$ until it contains less than a given number of points -- the bucket size $b$ or its extent is less than a minimal extent -- $e_{min}$. 
For memory efficient, octants without points are not created.
\begin{figure}[t]
	\setlength\abovecaptionskip{-0.1\baselineskip}
	\centering
	\includegraphics[width=0.4\textwidth]{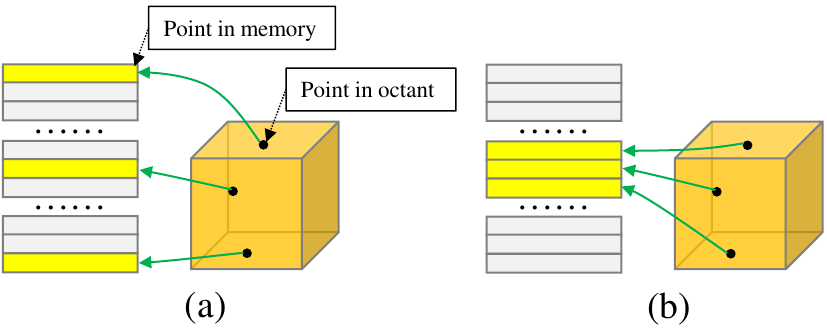}
	\caption{Illustration of locations of a octant's points  in memory. (a) scattered locations; (b) continual locations.}
	\label{fig:point_store}    
	\vspace{-0.4cm}
\end{figure}
In addition, we only keep indices and coordinates of points in each leaf octant and there is no points in non-leaf octants.
To enable extremely fast access to each point in each leaf octant, we propose a local spatially continuous storing strategy (as shown in Fig. \ref{fig:point_store}) which reallocates a segment of continuous memory for storing points information in the leaf octant after subdivision.
Furthermore, the reallocation facilitates box-wise delete and incremental update, as it allows operating on a segment of memory without influencing others.
% 局部连续空间存储策略 local spatially continuous storing strategy 可以和octree对比，这点对于后续的增量更新、下采样、box删除也很有帮助，看看是否需要在相关工作里说明一下

Based on above consideration, an octant $C_o$ of our \emph{i-Octree} contains the center $\boldsymbol{c}_o \in \mathbb{R}^3$, the extent  $e_o$, the points $\boldsymbol{P}_o$ storing coordinates and indices,  and a pointer pointing to the address of the first child octant.
The subscript ``o" is used to distinguish between different octants.
Particularly, let $C_r$ be the root octant, and $\boldsymbol{P}_r, e_r$, and  $\boldsymbol{c}_r$ are the points, the extent, and the center, respectively.

As for building an incremental octree, we firstly eliminate invalid points, calculate the axis-aligned bounding box of all valid points, and keep only indices and coordinates of valid points.
Then, starting at the root, the \emph{i-Octree}  recursively splits the axis-aligned bounding box at the center into eight cubes indexed by Morton codes\cite{Morton_codes} and subdivides all the points in current octant into each cube  according to their cube indices calculated.
% Z order 曲线
When a stopping criteria is satisfied, a leaf octant will be created and a segment of continuous memory will be allocated to store information in points of the leaf node.

\subsection{Dynamic Updates}\label{subsec:increupdates}
The dynamic updates include insertion of one or more points (i.e, incremental update) and delete of all points in an axis-aligned box (i.e., box-wise delete).
The insertion is integrated with down-sampling, which maintains the octree at a pre-determined resolution.

\subsubsection{Incremental Update}
When inserting new points, we have to consider the situation that some points may be  beyond the boundary of the axis-aligned bounding box of the original tree.
Once there are points out of the range of octree, we have to expand the bounding box by creating new root octant whose children contain current root octant.
\begin{figure}[t]
	\setlength\abovecaptionskip{-0.1\baselineskip}
	\centering
	\includegraphics[width=0.4\textwidth]{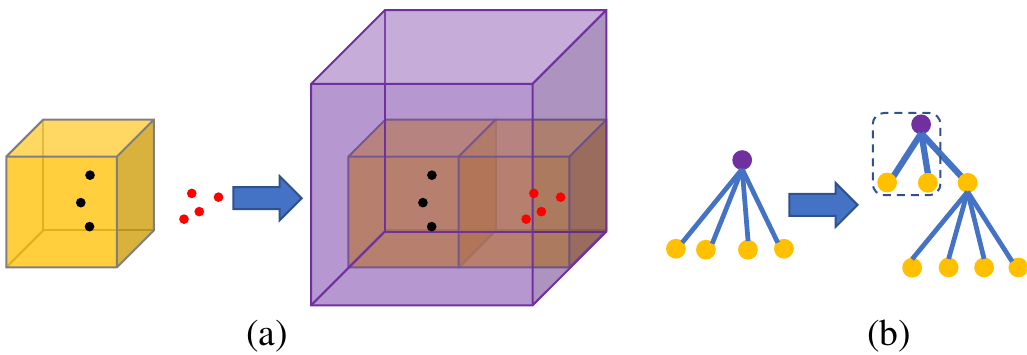}
	\caption{ Fig. (a) and (b) illustrate  insertion of new points (red) out of  the range to the \emph{i-Octree}. In (a), the left yellow cube is the root octant as well as the leaf octant with points (black) in the beginning. After inserting, the root octant becomes the light purple cube.  In (b), the purple node is the root octant and it is updated after inserting new octants (dashed black rectangle).}
		\vspace{-0.5cm}
	\label{fig:update_octree} 
\end{figure}
This process may be executed several times to ensure that all new points are within the range of the tree.
Then, new points are added to the expanded octree (see Fig. \ref{fig:update_octree}).

In consideration of efficient points queries in robotic applications, \emph{i-Octree} supports down-sampling that executes simultaneously with point insertion.
The down-sampling focuses on the new points and deletes the points that satisfy a certain condition: they are subdivided into a leaf octant whose extent is less than $2e_{min}$ and size is larger than $b/8$. 

%The algorithm of point insertion after expanding is detailed in {\bf Algorithm \ref{alg:update_octant}}.
The process of adding new points to an octant $C_o$ is similar to the construction of an octant.
If $C_o$ is a leaf node and it satisfies the subdivision criteria, all points (i.e., old points and new added points) in $C_o$ will be recursively subdivided into child octants.
If down-sampling is enabled, $e_{o}\le 2e_{min}$, and $|\boldsymbol{P}_{o}|>b/8$, new points will be deleted later instead of being added to $C_o$.
Otherwise, a segment of continuous memory will be allocated for the updated points.
If $C_o$ has child octants, the problem becomes assigning the newly added points to various children, and only the new points are to be subdivided.
This process is similar to the one mentioned above, except for the recursive updating of octants.

\subsubsection{Box-wise Delete}
In certain robotic applications, such as SLAM, only the points near the agent are required to estimate the state. 
Consequently, the points located far away from the agent in the \emph{i-Octree} are not essential and can be removed for efficiency concerns.

When it comes to removing unnecessary points in an axis-aligned cuboid, instead of directly searching for points within the given space and deleting them, the \emph{i-Octree} firstly checks whether the octants are inside the given box.
All octants inside the given box will be directly deleted without searching for points in them, which significantly reduces the deletion  time.
The deletion  of octants have no influence on others thanks to the local spatially continuous storing strategy.
\begin{figure}[t]
	\setlength\abovecaptionskip{-0.1\baselineskip}
	\centering
	\includegraphics[width=0.4\textwidth]{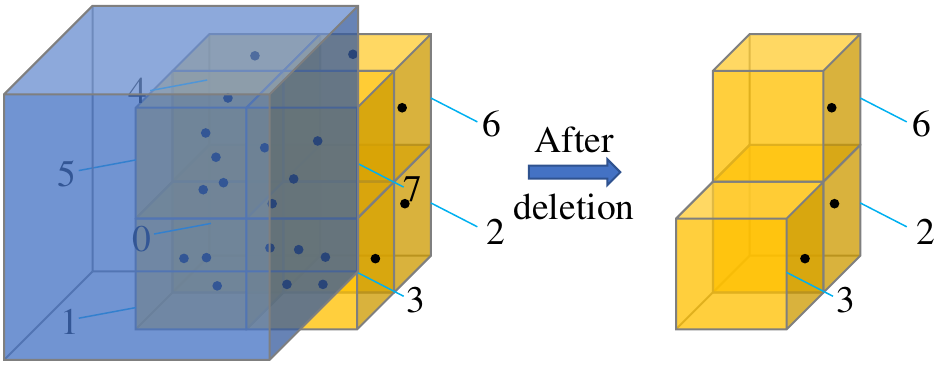}
	\caption{Illustration of box-wise delete. The blue box is the given box, and the numbers 0 to 7 (Morton codes)  are the indices of the octants. Octants, with indices 0, 1, 4, and 5 are directly deleted. Octant 7 is also deleted due to having no points.}
	\label{fig:box_delete}    
	\vspace{-0.4cm}
\end{figure}
For the leaf octants overlapping with the given box, we delete the points within the box and allocate a new segment of  memory for the remaining.
If a leaf octant contains no points after deletion, it will be deleted as shown in Fig. \ref{fig:box_delete}.

\subsection{K-Nearest Neighbors Search}\label{subsec:knn_search}
%(比如knn这里 你应该说summary你这里的功能和输入输出是什么, 现在是直接讲K-nearest neighbors search and radius neighbor search are essential for many robotic applications. Readers can refer to [17] for the octree with radius neighbor search, and in this paper we focus on k nearest neighbors search which is not enabled in [17]. 和文章的method部分没太大关系， 尤其第一句话这种 没有什么信息含量， 每一部分的行文应该是总分结构，好像每个part都有这个问题)
Using \emph{i-Octree}, we can retrieve $k$ nearest neighbors for an arbitrary query point $\boldsymbol{q}\in \mathbb{R}^3$.
The nearest search on the \emph{i-Octree} is an accurate nearest search \cite{friedman1975nearestsearch} instead of an approximate one as \cite{muja2009flannfast}.
We maintain a priority queue $\boldsymbol{h}$  with a maximal length of $k$ to store the k-nearest neighbors so far encountered and their distance to the query point $\boldsymbol{q}$.
The last element of $\boldsymbol{h}$ always has the largest distance regardless of pushing or popping.
The axis-aligned box of each octant is well utilized effectively  to accelerate  the nearest neighbor search using a “bounds-overlap-ball” test\cite{10.1145/355744.355745} and a proposed priority search order pre-computed according to the fixed indices of child octants.

Firstly, we recursively search down the \emph{i-Octree} from its root node until reach the leaf node closest to  $\boldsymbol{q}$. 
Then the distances from $\boldsymbol{q}$ to all points in the leaf node and corresponding indices will be pushed to priority queue $\boldsymbol{h}$.
All leaf nodes so far encountered will be searched before $\boldsymbol{h}$ is full.
If $\boldsymbol{h}$ is full and the search ball $\boldsymbol{S}(\boldsymbol{q}, d_{max})$ defined by  $\boldsymbol{q}$ and the largest distance $d_{max}$ in $\boldsymbol{h}$ is inside the axis-aligned box of current octant, the searching is over.
If a octant $C_k$  doesn't contain the search ball $\boldsymbol{S}(\boldsymbol{q}, d_{max})$, then one of the following three conditions must be satisfied:
\begin{equation}
	e_k - |{q}_x-{c}_{k,x}|  < d_{max},
\end{equation}
\begin{equation}
		e_k - |{q}_y-{c}_{k,y}|  < d_{max},
\end{equation}
\begin{equation}
		e_k - |{q}_z-{c}_{k,z}|  < d_{max},
\end{equation}
where $\boldsymbol{q} = ({q}_x, {q}_y, {q}_z)^T, \boldsymbol{c}_k = ({c}_{k,x}, {c}_{k,y}, {c}_{k,z})^T$. 
If none of the above conditions hold, the search ball is inside the octant.

We update $\boldsymbol{h}$ by investigating octants overlapping the search ball $\boldsymbol{S}(\boldsymbol{q}, d_{max})$, since only these could potentially contain points that are also inside the desired neighborhood.
We define the distance $d$  between  $\boldsymbol{q}$ and $C_k$ as below:
\begin{equation}
	d = \rVert  \sigma(|\boldsymbol{q}-\boldsymbol{c}_o|-\boldsymbol{1}e_o) \lVert_2,
\end{equation}
where $\boldsymbol{1}=(1,1,1)^T$ and $\sigma(x)=x$  if $x>0$, otherwise $\sigma(x)=0$.
$d<d_{max}$ indicates that $C_k$ overlaps $\boldsymbol{S}(\boldsymbol{q}, d_{max})$.
In order to speed up the process, we sort the candidate child octants of $C_o$ according to their distances to $C_k$ and get 8 different sequences in $\boldsymbol{I}_{order}$.
Such that the closer octants are earlier to be searched and the search reaches its end early.

\subsection{Radius Neighbors Search}\label{subsec:radius_search}
For an arbitrary query point $\boldsymbol{q}\in \mathbb{R}^3$ and radius $r$,  the radius neighbors search method finds every point $\boldsymbol{p}$ satisfying $\lVert \boldsymbol{p} - \boldsymbol{q}\rVert_2<r$.
The process is similar to KNN search except for a fixed radius and an unlimited $k$.
We adopt the pruning strategy proposed by Behley et al. \cite{Behley_et_2015_octree} with improvements to reduce computation cost.
Before test whether $\boldsymbol{S}(\boldsymbol{q}, r)$ completely contains an octant $C_k$, we make a simple test whether $r^2$ is larger than $3e_k^2$.
The simple test is the necessary condition for  $\boldsymbol{S}(\boldsymbol{q}, r)$ containing $C_k$ with small cost.
Besides, we try to avoid the extraction of a root in the algorithm.
These two tricks together with our local spatially continuous storing strategy play key roles in accelerating the searching.    

\section{Application Experiments}\label{sec:experiments}
We compare our \emph{i-Octree} to publicly available implementations of static k-d trees (i.e., k-d tree used in Point Cloud Library (PCL)) , incremental k-d tree (i.e., \emph{ikd-Tree}), and PCL octree.
We conduct the experiments on randomized data and various open real-world datasets.
We first evaluate our \emph{i-Octree} against PCL octree and the state-of-the-art incremental k-d tree, i.e., \emph{ikd-Tree}, for tree construction, point insertion, KNN search, radius neighbors search, and box-wise delete on randomized three dimensional (3D) point data of different size. 
Then, we validate the \emph{i-Octree} in actual robotic applications on real-world dataset by  replacing the static k-d tree in the LiDAR-based SLAM  without any refinement and evaluate the time performance and accuracy.
All experiments are performed on a PC with Intel i7-13700K CPU at 3.40GHz. 

\subsection{Randomized Data Experiments}
The efficiency of our \emph{i-Octree} is fully investigated by the following experiments on randomized incremental data. 

\subsubsection{Performance Comparison}
In this experiment, we investigate the time performance of three  implementations of dynamic data structure (i.e., the \emph{i-Octree}, the \emph{ikd-Tree}, and the PCL octree).
Both \emph{ikd-Tree} and PCL octree are state-of-the-art implementations with high efficiency and support point insertion, which is the key consideration why we choose them to be compared with our \emph{i-Octree} for a fair comparison.
In this experiment, 200,000 points are generated randomly in a $10m\times10m\times10m$ space (i.e., the workspace) to build the trees.
%The cubic length $l_c$ has four values, i.e. $0.1m$, $1m$, $5m$ and $10m$, for different levels of point cloud density.
Then 100 test operations are conducted.
In each test operation, 2000 new points randomly sampled in the workspace are inserted to the trees.
Besides, we perform KNN search with $k=5$ on each of 200 new points randomly sampled and radius neighbors search with $r=0.3m$ on  each of another 200 ones.
Down-sampling is not enabled on \emph{i-Octree} and \emph{ikd-Tree} in this experiment.
We fix $e_{min} = 0.01m$ for \emph{i-Octree} and PCL octree and $b=32$ for \emph{i-Octree} in the following experiments.
We record the time for tree construction and point insertion, the total time of KNN search for 200 points, and the total time of  radius neighbors search for another 200 points at each step.

\begin{figure}[t]
	\setlength\abovecaptionskip{-0.4cm}
	\centering
	\includegraphics[width=0.485\textwidth]{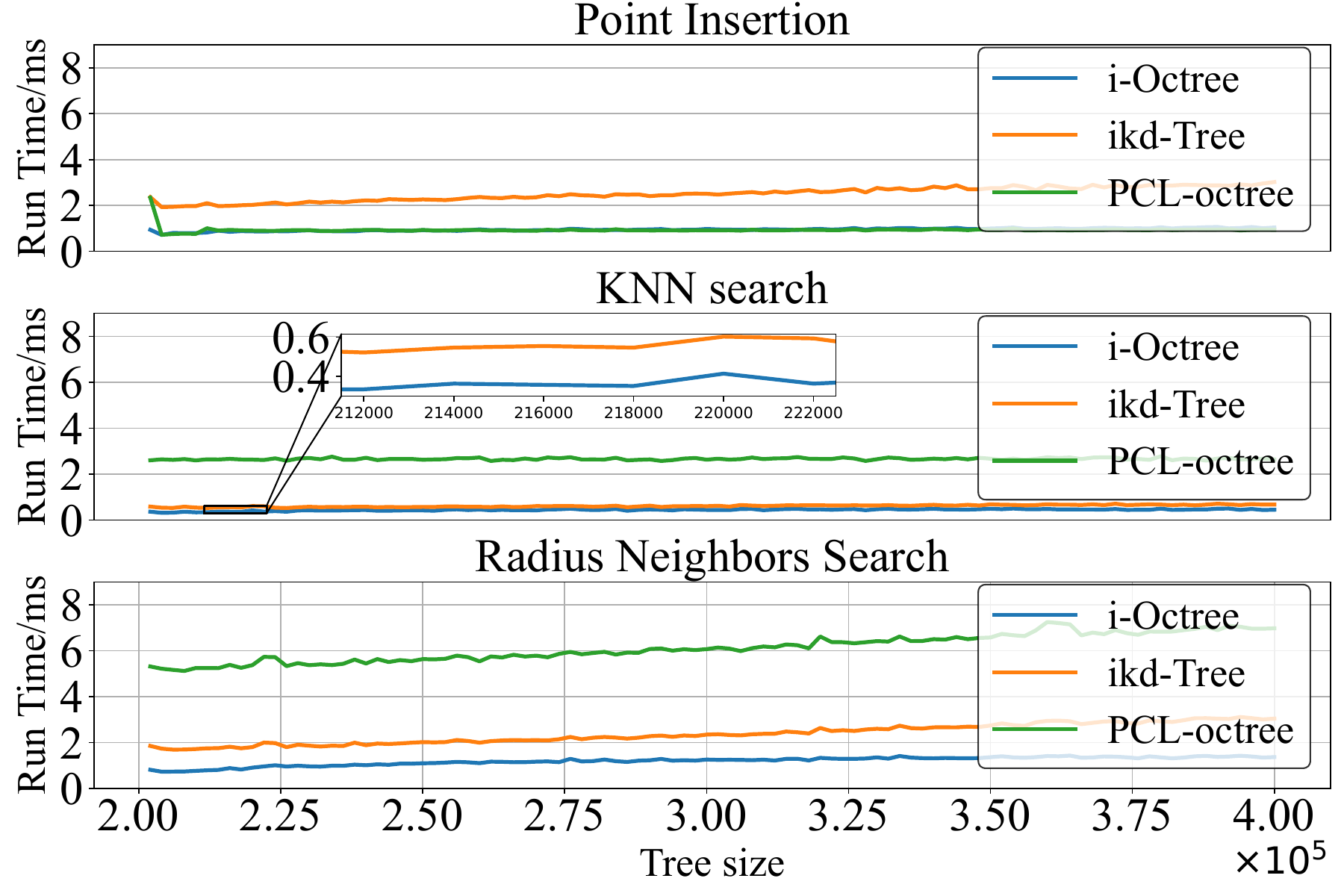} % 
	\caption{Dynamic data structure comparison over different tree size.}
	\label{fig:random}    
	\vspace{-1.0cm}
\end{figure}

The dynamic data structure comparison over different tree size is shown in Fig. \ref{fig:random}.
The Table \ref{tab:rand_aver} shows the comparison of average time consumption and peak memory usage.
When the tree size increases from 200,000 to 400,000, the time for point insertion (without down-sampling) of \emph{i-Octree} and PCL octree remains stable at 0.8$ms$ while that for the \emph{ikd-Tree} is 3 times larger and grows linearly with the tree size.
%\addtolength{\tabcolsep}{0.5em} 
\begin{table}[h]
	\caption{The Comparison of Average Time Consumption and Peak Memory Usage}
	\label{tab:rand_aver}
	\centering
	\begin{threeparttable}
		\setlength{\tabcolsep}{0.6mm}{ % 缩小列宽
			\begin{tabular}{cccc} 
				\toprule[1pt] % 表格顶部的粗线
%				\multirow{2}*{}	 & \multicolumn{3}{c}{$l_c = 10m$}   \\ 	\cmidrule(l){2-4}
					& \emph{i-Octree}	&  \emph{ikd-Tree}  & PCL octree 	\\	\midrule
				Construction[ms]\tnote{1}  	  & \textbf{16.63}     & 46.26     &   52.66   \\ 
				Point Insertion[ms] 				 & \textbf{0.83}       &  2.45      &   0.85     \\
				KNN Search[ms]					& \textbf{0.41}       &  0.59      &   2.53   \\
				Radius Neighbors Search[ms] 	& \textbf{0.81}   	 &   1.85     &   3.96 	    \\ 
				Peak Memory Usage[Mb]   & \textbf{21.53}   	 &   62.91    &  136.82 	    \\
				\bottomrule[1pt]
			\end{tabular}
		}
		\begin{tablenotes}
			\footnotesize
			\item[1] The construction of the tree is conducted only once  at the first time.
		\end{tablenotes}
	\end{threeparttable}
	\vspace{-0.4cm}
\end{table}
%\addtolength{\tabcolsep}{-0.5em} 
% 这里可以适当删减一些
For KNN search, our method demonstrates a performance that is more than twice as fast as the PCL octree and achieves a 20\% reduction in runtime compared to the \emph{ikd-Tree}.
Regarding the execution time for radius neighbors search, the \emph{i-Octree} demonstrates a performance that is more than twice as fast as both the \emph{ikd-Tree} and the PCL octree.
Furthermore, the construction time of the \emph{i-Octree} is less than 36\% compared to the \emph{ikd-Tree}. 
Additionally, the peak memory usage of the \emph{i-Octree} is the lowest, being less than 35\% of that observed for the \emph{ikd-Tree}.

\subsubsection{Box-wise Delete}
The \emph{ikd-Tree} is chosen to be compared with the \emph{i-Octree} since both support the box-wise delete.
This experiment investigates the time performance of deleting all points in an axis-aligned box and the time performance of KNN serach and radius neighbors search after delete. 
In the experiment, we sample 400,000 points randomly in a $10m\times10m\times10m$ space (i.e., the workspace) to initialize the incremental octree. 
Then 1,00 test operations are conducted on the trees. 
%	mean knn:  [0.17353011 0.39710988]
%	mean radius:  [0.44873002 1.31936058]
\begin{figure}[t]
	\centering
	\setlength\abovecaptionskip{-0.4cm}
	\includegraphics[width=0.485\textwidth]{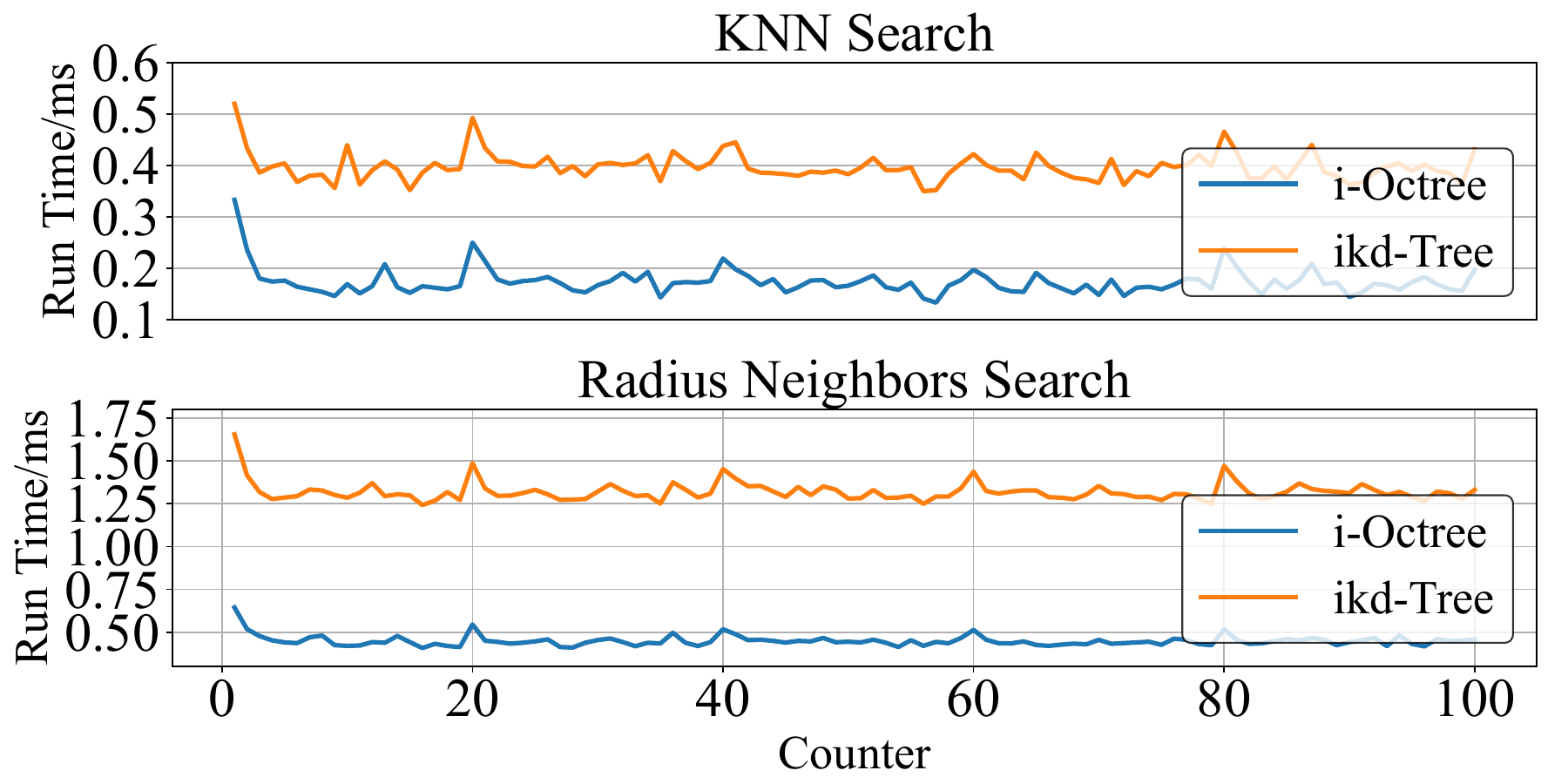}  % 
	\caption{The comparison of time consumption.}
	\label{fig:rand_box_delete}    
	\vspace{-1.0cm}
\end{figure}
In each test operation, we randomly sample 200 points in the workspace for KNN search and another 200 points for radius neighbors search.
For every 20 test operations, one axis-aligned box is sampled in the workspace with side length of $1.0m$ and points contained in the box are deleted from the trees.
We record the tree size after each delete, the time of box-wise delete,  the total time of KNN search for 200 points, and the total time of  radius neighbors search for another 200 points at each step.

The time of box-wise delete and the tree size are shown in Table \ref{tab:rand_box_wise}.
%\addtolength{\tabcolsep}{0.5em} 
\begin{table}[h]
	\caption{The Time Consumption of Box-wise Delete and Tree Size}
	\label{tab:rand_box_wise}
	\centering
	%	\usepackage{booktabs}    %导言区
	% 列宽调整 https://blog.csdn.net/weixin_41271939/article/details/121993833  \makebox
	%	\toprule命令：表格顶部的粗线。
	%	\midrule命令：表格中间的细分隔线。
	%	\bottomrule命令：表格底部的粗线。
	\begin{threeparttable}
		\setlength{\tabcolsep}{2.5mm}{
			\begin{tabular}{cccccc} 
				\toprule[1pt] % 表格顶部的粗线
				\multirow{2}*{Counter}	 & \multicolumn{2}{c}{Box-wise Delete[ms]} & \multicolumn{3}{c}{Tree Size}   \\ 	\cmidrule(l){2-3}  \cmidrule(l){4-6}
				& \emph{i-Octree}              & \emph{ikd-Tree} & GT\tnote{1}& \emph{i-Octree}   &  \emph{ikd-Tree}       \\  		\midrule
				0       & --                        & --           & 400,000      & 400,000   & 400,000    \\
				20     & \textbf{0.017}    & 0.222     & 399,614     & 399,614   & 399,623  \\
				40     & \textbf{0.013}    & 0.343     & 399,205     & 399,205   & 399,236   \\
				60     & \textbf{0.017}    & 0.250     &398,783      & 398,783   & 398,826   \\
				80     & \textbf{0.016}    & 0.320     & 398,375     & 398,375   & 398,437   \\
				100   & \textbf{0.006}    & 0.126     & 398,185     & 398,185   & 398,247   \\
				\bottomrule[1pt]
			\end{tabular}
		}
		\begin{tablenotes}
			\footnotesize
			%			\item[1] The time consumption of box-wise delete.
			\item[1] GT denotes the exact size of the point cloud at each test.
		\end{tablenotes}
	\end{threeparttable}
\vspace{-0.2cm}
\end{table}
%\addtolength{\tabcolsep}{-0.5em} 
The run time performance of \emph{i-Octree} and \emph{ikd-Tree} on KNN search and radius neighbors search is shown in Fig. \ref{fig:rand_box_delete}.
In this experiment, it appears that the box-wise delete operation on the \emph{ikd-Tree} does not remove all points within the given box as shown in Table \ref{tab:rand_box_wise}.
The results show that our \emph{i-Octree} runs over 10 times faster than the \emph{ikd-Tree} on the box-wise delete and approximately 3 times faster on the radius neighbors search.
Besides, the \emph{i-Octree}  reduces the run time of KNN search by 58\% over the \emph{ikd-Tree}.
%	mean knn:  [0.17353011 0.39710988]
%	mean radius:  [0.44873002 1.31936058]

\subsection{Real-world Data Experiments}
% 将\emph{i-Octree} 替代已有算法中的静态ke-tree，检验算法性能。
% 测试场景说清楚，做了三个数据集上的实验，达到了什么效果
We test our developed \emph{i-Octree} in  actual robotic applications (e.g., LiDAR-inertial SLAM\cite{FAST-LIO2, shan2020lio_sam,xu2020fastlio, liliom} and  pure LiDAR SLAM\cite{wang2021floam, shan2018lego,zhang2014loam}).
In these experiments, we directly replace the static k-d tree in the LiDAR-based SLAM by our proposed \emph{i-Octree} without any refinement and evaluate the time performance and accuracy.
To ensure a fair and complete comparison, we test on three datasets: M2DGR\cite{yin2021m2dgr}, Newer College Dataset\cite{zhang2021multicamera} and NCLT\cite{NCLT_ncarlevaris_2015a} with different sensor setups.
The M2DGR is a low-beam LiDAR data, the Newer College Dataset is a high-beam LiDAR data, and the NCLT is a long-term dataset.
They all have ground truth trajectories such that we can evaluate the accuracy of the LiDAR trajectories using absolute trajectory error (ATE)\cite{sturm2012benchmark}.

For the sake of simplicity, we will focus on three representative LiDAR-based SLAM algorithms: FAST-LIO2\cite{FAST-LIO2} and LIO-SAM\cite{shan2020lio_sam} that combine LiDAR and inertial measurements (LiDAR-inertial SLAM), and FLOAM\cite{wang2021floam} that relies solely on LiDAR data (pure LiDAR SLAM).
LIO-SAM and  FLOAM are both the state-of-the-art algorithms and uses the static k-d tree for the nearest points search, which  is crucial to match a point in a new LiDAR scan to its correspondences in the map (or the previous scan).
For the LIO-SAM, we choose the LIO\_SAM\_6AXIS\footnote[1]{\url{https://github.com/JokerJohn/LIO_SAM_6AXIS}}, which is modified from LIO-SAM to support a wider range of sensors (e.g., 6-axis IMU and low-cost GNSS) without accuracy loss.
Besides, the loop closure module of LIO\_SAM\_6AXIS is deactivated.
The parameters setups, as shown in Table \ref{tab:param},  for every sequence in a dataset are the same.
Both algorithms build two independent maps: surface map and edge map, where two k-d trees or \emph{i-Octree} will be built.
The SurfRes and EdgeRes determine the minimal extent of the two \emph{i-Octree} structures,  respectively.
%Moreover, the performance comparison of \emph{i-Octree} and \emph{ikd-Tree} integrated into FAST-LIO2 is tested on Newer College Dataset and NCLT.
Moreover, the performance of \emph{i-Octree} and \emph{ikd-Tree}, once integrated into FAST-LIO2, is evaluated using the Newer College Dataset.
%\addtolength{\tabcolsep}{0.5em} 
\begin{table}[h]
	\caption{The Parameters Setup for each dataset}
	\label{tab:param}
	\centering
	%	\usepackage{booktabs}    %导言区
	% 列宽调整 https://blog.csdn.net/weixin_41271939/article/details/121993833  \makebox
	%	\toprule命令：表格顶部的粗线。
	%	\midrule命令：表格中间的细分隔线。
	%	\bottomrule命令：表格底部的粗线。
	\begin{threeparttable}
		\setlength{\tabcolsep}{0.8mm}{
			\begin{tabular}{ccccccc} 
				\toprule[1pt] % 表格顶部的粗线
				\multirow{2}*{Dataset}	 & \multicolumn{3}{c}{FLOAM} & \multicolumn{3}{c}{LIO\_SAM\_6AXIS}   \\ 	\cmidrule(l){2-4}  \cmidrule(l){5-7}
				& SurfRes     & EdgeRes  & MinRange & SurfRes & EdgeRes   &  MinRange       \\  		\midrule
				M2DGR       
				& 0.1m        & 0.05m   & 0.2m   & 0.4m      & 0.2m  & 1.0m    \\ \addlinespace
				\begin{tabular}[c]{@{}c@{}}Newer\\      College\end{tabular}         
				& 0.2m        & 0.1m   & 0.2m   & 0.4m      & 0.2m  & 1.0m    \\ \addlinespace
				NCLT        
				& 0.4m        & 0.2m   & 0.2m   & 0.4m      & 0.2m  & 1.0m    \\
				\bottomrule[1pt]
			\end{tabular}
		}
		\begin{tablenotes}
			\footnotesize
			%			\item[1] The time consumption of box-wise delete.
			\item[*] SurfRes and  EdgeRes denote the map resolution of surface map and edge map,  respectively;  MinRange means that any points within the LiDAR  range defined by  MinRange will not  be considered.
		\end{tablenotes}
	\end{threeparttable}
		\vspace{-0.4cm}
\end{table}
%\addtolength{\tabcolsep}{-0.5em} 

% 低线束
\subsubsection{Low-beam LiDAR Data}
The first dataset is M2DGR, which is a  large-scale dataset collected by an unmanned ground vehicle (UGV) with a full sensor-suite including a Velodyne VLP-32C LiDAR sampled at 10 $Hz$, a 9-axis Handsfree A9 inertial measurement unit (IMU) sampled at 150 $Hz$,  and other sensors.
The dataset comprises 36 sequences  captured in diverse scenarios including both indoor and outdoor environments on the university campus.
The ground truth trajectories of the outdoor sequences are obtained by the Xsens MTi 680G GNSS-RTK suite while for the indoor environment, a motion-capture system named Vicon Vero 2.2 whose localization accuracy is 1$mm$  is used to collect the ground truth.
We choose several sequences from this dataset and the detailed information is summarized in Table \ref{tab:new_college_time_rmse}.
  
\begin{figure}[t]
	\setlength\abovecaptionskip{-0.4cm}        
	\centering
	\includegraphics[width=0.485\textwidth]{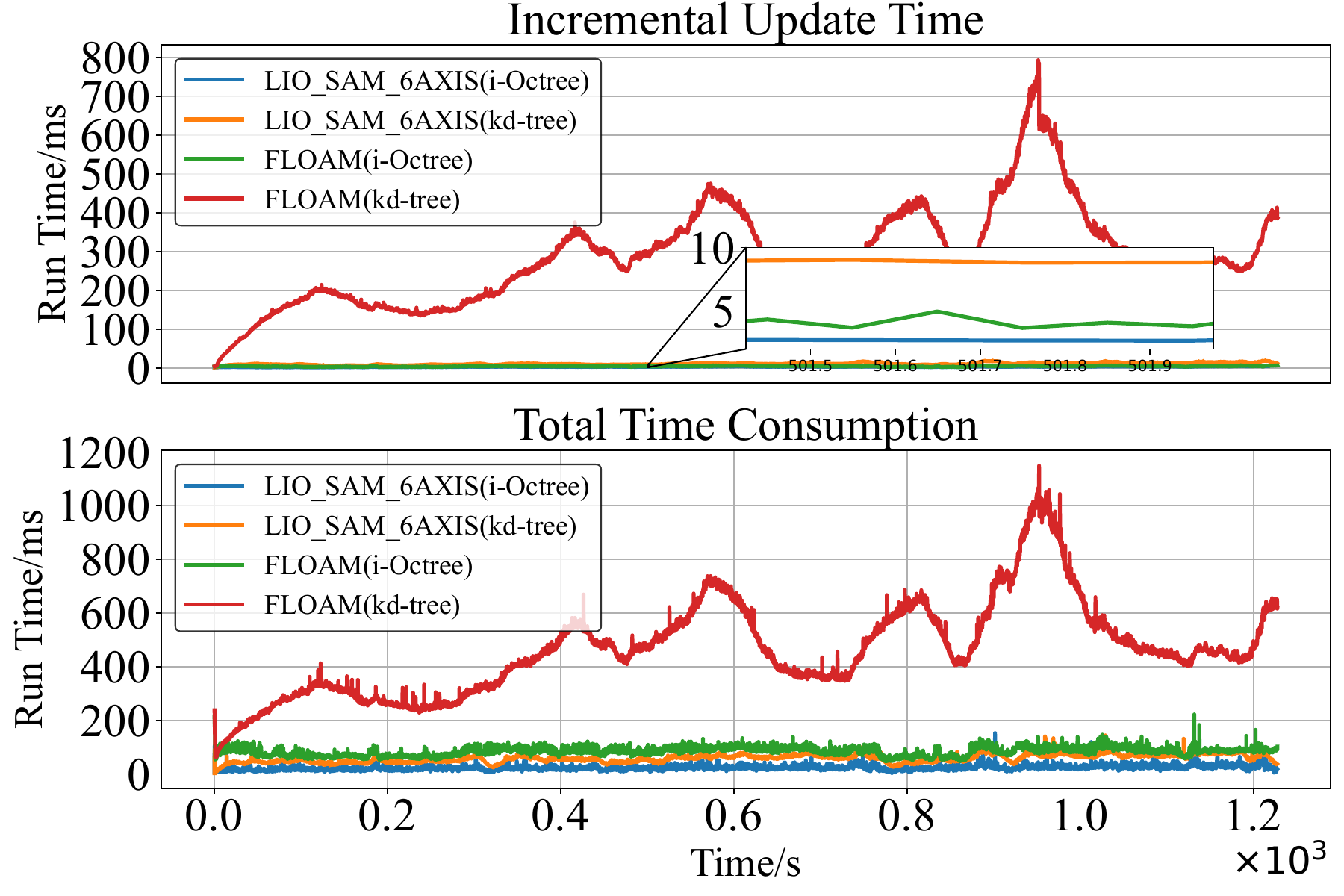} % 
	\caption{The Comparison of time consumption on the \emph{street\_02} sequence.}
	\label{fig:m2dgr_street_02}
	\vspace{-1.4cm}
\end{figure}

The maximal and average time consumption of incremental update for each scan are listed in Table \ref{tab:m2dgr_time_result}.
%\addtolength{\tabcolsep}{0.5em} 
\begin{table}[t]
	\caption{The Comparison of Maximal and Average Time Consumption per Scan on Incremental Update}
	\label{tab:m2dgr_time_result}
	\centering
	%	\usepackage{booktabs}    %导言区
	% 列宽调整 https://blog.csdn.net/weixin_41271939/article/details/121993833  \makebox
	%	\toprule命令：表格顶部的粗线。
	%	\midrule命令：表格中间的细分隔线。
	%	\bottomrule命令：表格底部的粗线。
	\begin{threeparttable}
		\setlength{\tabcolsep}{1.1mm}{
			\begin{tabular}{ccccccccc} 
				\toprule[1pt] % 表格顶部的粗线
				\multirow{5}*{\begin{tabular}[c]{@{}c@{}}Sequence\\      Name\end{tabular}  }	 
				& \multicolumn{4}{c}{FLOAM[ms]} & \multicolumn{4}{c}{LIO\_SAM\_6AXIS[ms]}   \\ 	\cmidrule(l){2-5}  \cmidrule(l){6-9}
				& \multicolumn{2}{c}{Incre\_max}  & \multicolumn{2}{c}{Incre\_mean}  
				& \multicolumn{2}{c}{Incre\_max}  & \multicolumn{2}{c}{Incre\_mean}     \\  		 	
				\cmidrule(l){2-3}  \cmidrule(l){4-5} 	\cmidrule(l){6-7}  \cmidrule(l){8-9}
				& ioct  & orig   & ioct  & orig   
				& ioct  & orig   & ioct  & orig          \\           \midrule
				\emph{gate\_01}            	   & \textbf{7.20} & 969.98 & \textbf{5.57} & 354.73  & \textbf{5.16} & 33.99 & \textbf{2.82} & 17.56      \\
				\emph{walk\_01}   			   & \textbf{6.92} & 271.49 & \textbf{5.17} & 141.08   & \textbf{4.66} & 15.39 & \textbf{2.53} & 11.18         \\
				\emph{street\_02} 		 		& \textbf{7.07} & 793.63 & \textbf{4.20} & 286.17    & \textbf{6.95} & 19.88 & \textbf{2.94} & 10.84   \\
				\emph{street\_03} 				& \textbf{9.22} & 1983.50 & \textbf{6.52} & 893.20    & \textbf{7.27} & 30.05 & \textbf{3.58} & 19.57                \\
				\emph{room\_01}     		 & \textbf{2.22} & 17.41 & \textbf{0.64} & 5.67    & \textbf{0.42} & 1.70 & \textbf{0.29} & 0.48       \\
				\emph{room\_02}     		 & \textbf{3.00} & 24.00 & \textbf{0.65} & 6.30    & \textbf{0.43} & 0.73 & \textbf{0.28} & 0.46  		             \\
				\emph{room\_dark\_01}    & \textbf{2.48} & 13.57 & \textbf{0.71} & 6.13    & \textbf{0.52} & 0.74 & \textbf{0.29} & 0.44            \\
				\bottomrule[1pt]
			\end{tabular}
		}
		\begin{tablenotes}
			\footnotesize
			\item[*] Note: Incre\_max  and Incre\_mean denote  the maximal and average incremental update time of each scan respectively;  ioct means \emph{i-Octree} and orig means the static k-d tree.
		\end{tablenotes}
	\end{threeparttable}
\vspace{-0.4cm}
\end{table}
%\addtolength{\tabcolsep}{-0.5em} 
For the FLOAM running in the outdoor environment, the size of map  grows gradually making it computational costly when using the static k-d tree.
The rebuilding process may cost a lot when the size of map is large, which deteriorates the real-time performance sharply.
\begin{table*}[!htb]
	\caption{The Comparison of Average Time Consumption per Scan and Absolute Translational Errors (RMSE, meters) }
	\label{tab:new_college_time_rmse}
	\centering
	%	\usepackage{booktabs}    %导言区
	% 列宽调整 https://blog.csdn.net/weixin_41271939/article/details/121993833  \makebox
	%	\toprule命令：表格顶部的粗线。
	%	\midrule命令：表格中间的细分隔线。
	%	\bottomrule命令：表格底部的粗线。
	\begin{threeparttable}
		\setlength{\tabcolsep}{1.1mm}{
			\begin{tabular}{cccccccccccc} 
				\toprule[1pt] % 表格顶部的粗线
				\multirow{5}*{\begin{tabular}[c]{@{}c@{}}Dataset\\      Name\end{tabular}  }		& \multirow{5}*{\begin{tabular}[c]{@{}c@{}}Sequence\\      Name\end{tabular}  }	 
				& \multicolumn{2}{c}{} & \multicolumn{4}{c}{Average Time Consumption per Scan[ms]} & \multicolumn{4}{c}{RMSE[m]}   \\   \cmidrule(l){5-8} \cmidrule(l){9-12}
				&	& \multicolumn{2}{c}{Attributes}  & \multicolumn{2}{c}{FLOAM}  & \multicolumn{2}{c}{LIO\_SAM\_6AXIS} & \multicolumn{2}{c}{FLOAM}  & \multicolumn{2}{c}{LIO\_SAM\_6AXIS} \\ \cmidrule(l){3-4}  \cmidrule(l){5-6} \cmidrule(l){7-8} \cmidrule(l){9-10} \cmidrule(l){11-12}
				&	& Duration &  Length  & \emph{i-Octree}  	& static k-d tree & \emph{i-Octree}  	& static k-d tree & \emph{i-Octree}  	& static k-d tree & \emph{i-Octree}  	& static k-d tree     \\  		  \midrule	
				\multirow{8}*{M2DGR  }	
				&	\emph{gate\_01}            & 172s           & -      & 101.70 & 570.62     & \textbf{40.69}  & 100.50     & 8.667 & 9.206 & \textbf{0.167} & 0.340      \\   
				&	\emph{walk\_01}   & 291s   & -       & 91.21 & 273.72  & \textbf{18.98}  & 52.42  & 0.108 & 0.087 &\textbf{0.078} & 0.079    \\
				&	\emph{street\_02} &  1227s  & -   & 82.85 & 462.21  & \textbf{24.26}  & 58.0  & 133.602 & 80.935 & \textbf{2.296} & 2.942      \\
				&	\emph{street\_03} &  354s  & -  & 117.55 & 1318.47      & \textbf{39.11}  & 114.55    & 15.781 & 16.923 & \textbf{0.137} & 0.138     \\
				&	\emph{room\_01}     & 72s   & -   & 22.42 & 25.91        & \textbf{3.11}  & 12.32      & \textbf{0.134} & 0.135 & 0.137 & 0.209     \\
				&	\emph{room\_02}     & 75s    & -    & 22.52 & 26.60         & \textbf{3.02}  & 6.32      & 0.128 & 0.129 & 0.123 & \textbf{0.121}     \\
				&	\emph{room\_dark\_01}     & 111s    & -  &   22.85 & 26.16     & \textbf{2.70}  & 5.73     & \textbf{0.146} & 0.148 & 0.151 & 0.155      \\ \midrule
				\multirow{5}*{\begin{tabular}[c]{@{}c@{}}Newer\\      College\end{tabular}  }	
				&	\emph{Quad-Easy}            & 198s          & 247m         	& 73.87 & 225.31 & \textbf{38.03}  & 100.73  & 0.795 & 0.399 & 0.083 & \textbf{0.079}      \\
				&	\emph{Quad-Medium}     & 190s    		& 260m  	   & 58.19  & 430.78 & \textbf{43.02}  & 96.86  & 15.626 & 18.358 & 0.100 & \textbf{0.095}      \\
				&	\emph{Quad-Hard}     & 187s    		& 234m  		 	& 44.86  & 300.38 & \textbf{32.85}  & 70.35  & 17.133 & 7.326 & \textbf{0.130}  & 0.152     \\
				&	\emph{Cloister}   				& 278s   		& 429m   		& 41.06  & 156.80 & \textbf{20.07}  & 53.77   & 15.100 & 7.685 & \textbf{0.115}  & 0.169  \\
				&	\emph{Maths-Easy} 			&  216s  		& 264m  	& 62.53  & 180.89 & \textbf{31.34}  & 74.46  	& 1.026 & 0.260 & \textbf{0.088}  & 0.125     \\
				&	\emph{Underground-Easy} 	&  141s  		& 162m & 28.16  & 41.27 & \textbf{9.37}  & 30.85  & 0.143 & 0.078 & 0.077  & \textbf{0.073}     \\  \midrule
				NCLT
				&	\emph{2012-01-08} 			&5633s &6.4km	& \textbf{21.88} & 70.62 & 24.65 & 53.54     	& 187.588 & 62.188 & \textbf{1.890} & 9.291        \\
				\bottomrule[1pt]
			\end{tabular}
		}
		%		\begin{tablenotes}
		%			\footnotesize
		%			\item[*] Note: Incre\_max  and Incre\_mean denote  the maximal and average incremental update time of each scan respectively; Total\_max  and Total\_mean denote the maximal and average  total time consumption for each scan; ioct means \emph{i-Octree} and orig means the static k-d tree.
		%		\end{tablenotes}
	\end{threeparttable}
	\vspace{-0.3cm}
\end{table*}
The time for  rebuilding  the entire k-d tree fills a large percentage of the total time per scan, as shown in the Fig. \ref{fig:m2dgr_street_02}.
The time for  rebuilding  the entire k-d tree fills a large percentage of the total time per scan.
After replacing the k-d tree by \emph{i-Octree}, the incremental time reduces  to less than 10$ms$ and the real-time performance is approximately guaranteed.
The FLOAM with \emph{i-Octree} runs over 5 times faster than the original one,  with sometimes a slight accuracy loss as shown in Table \ref{tab:new_college_time_rmse}.
The LIO\_SAM\_6AXIS with \emph{i-Octree} runs over 2 times faster than original one on almost all sequences and the accuracy is usually improved.
Besides, the \emph{i-Octree} reduces the peak time which degrades the real-time ability.

% 大数据量
\subsubsection{High-beam LiDAR Data}
The second dataset is the Newer College Dataset, which is a multi-camera LiDAR inertial dataset of 4.5 $km$ walking distance.
The dataset is collected by a handheld device equipped with an Ouster OS0-128 LiDAR sampled at 10 $Hz$ with embedded IMU sampled at 100 $Hz$.
Precise centimetre accurate ground truth is provided and calculated by registering each undistorted LiDAR frame to a prior map scanned by 3D imaging laser scanner, Leica BLK360.
Table \ref{tab:new_college_time_rmse} shows the 6 sequences selected from the dataset.
These sequences contain small and narrow passages, large scale open spaces, as well as vegetated areas.
Besides, challenging situations such as aggressive motion are presented.

In this test, we record average total time consumption of each scan and calculate the RMSE of the absolute translational  errors, as shown in Table \ref{tab:new_college_time_rmse}.
We find that the \emph{i-Octree} shows promising compatibility with LIO\_SAM\_6AXIS since it not only reduces the time consumption a lot but also improves the accuracy.
For the LOAM, the \emph{i-Octree} significantly accelerates the processing speed per scan, but sometimes at a cost of degrading accuracy, as it directly replaces the static k-d tree without any refinement.
%\addtolength{\tabcolsep}{0.5em} 
\begin{table}[h]
	\vspace{-0.1cm}
	\caption{Performance Comparison of \emph{i-Octree} and \emph{ikd-Tree} }
	\label{tab:fast_lio2_comp}
	\centering
	\begin{threeparttable}
		\setlength{\tabcolsep}{0.5mm}{
			\begin{tabular}{ccccccc} 
				\toprule[1pt] % 表格顶部的粗线
				\multirow{3}*{\begin{tabular}[c]{@{}c@{}}Sequence\\      Name\end{tabular}  }	 
				& \multicolumn{2}{c}{AveTim[ms]} & \multicolumn{2}{c}{RMSE[m]} & \multicolumn{2}{c}{PeaMem[Mb]}   \\ 	\cmidrule(l){2-3}  \cmidrule(l){4-5} \cmidrule(l){6-7}
				& \emph{i-Octree}  & \emph{ikd-Tree}    & \emph{i-Octree}  &   \emph{ikd-Tree}   & \emph{i-Octree}  &   \emph{ikd-Tree}          \\           \midrule
				\emph{Quad-Easy}            	& \textbf{14.58} & 19.70 &\textbf{0.069} & 0.071    & \textbf{242.06}& 252.26   \\
				\emph{Quad-Medium}     		 & \textbf{14.24} & 16.21 & 0.059 & 0.059  & \textbf{239.04} & 259.13    \\
				\emph{Quad-Hard}    		 	& \textbf{10.35} & 12.08 & 0.051& \textbf{0.047}    & \textbf{240.58} & 257.86    \\
				\emph{Cloister}   					&  \textbf{7.30}  & 9.08 & \textbf{0.053} & 0.063 & \textbf{255.93} & 261.62  \\
				\emph{Maths-Easy} 				&  \textbf{12.45} & 16.41  & 0.093  & 0.093   &  \textbf{252.77} & 278.04   \\
				\emph{Underground-Easy}  &  \textbf{4.82} & 5.96 &  \textbf{0.029}  & 0.036&  \textbf{235.12} & 239.61   \\  
				\bottomrule[1pt]
			\end{tabular}
		}
				\begin{tablenotes}
					\footnotesize
					\item[*] Note: AveTim denotes the average time required for updating the tree and executing KNN search for each scan ; PeaMem means the peak memory usage of ROS node.
				\end{tablenotes}
	\end{threeparttable}
	%	\vspace{-0.3cm}
\end{table}
% 26 12 14  19 24  19 ==》 19
Furthermore, the performance comparison between \emph{i-Octree} and \emph{ikd-Tree}, as integrated into FAST-LIO2 without parallelization, is evaluated using this dataset, as presented in Table~\ref{tab:fast_lio2_comp}.
In this experiment, we set $e_{min} = 0.5m$ for both \emph{i-Octree} and \emph{ikd-Tree}, and $b=1$ specifically for \emph{i-Octree}. 
The other \emph{ikd-Tree} parameters are kept at their default settings.
%The \emph{i-Octree} outperforms the \emph{ikd-Tree} in terms of memory efficiency and speed by achieving, on average, a 19\% reduction in runtime  without compromising accuracy, aligning with results from experiments conducted on randomized data.
The \emph{i-Octree} method exhibits superior memory efficiency and speed compared to the \emph{ikd-Tree}, achieving an average reduction in runtime of 19\% without sacrificing accuracy, consistent with findings from experiments on randomized datasets.

% 长时间
\subsubsection{Long-term LiDAR Data}
Finally, we test on a long sequence called \emph{2012-01-08} form a  large scale dataset named NCLT captured by a Velodyne HDL-32E LiDAR sampled at 10 $Hz$, a Microstrain 3DM-GX3-45 IMU sampled at 100 $Hz$, and other sensors.
The length of the sequence is 6.4$km$ and the duration is 5633s.
The average run time of each scan and the absolute translational errors are added to Table \ref{tab:new_college_time_rmse}.
The FLOAM drifts severely on the large scale sequence while LIO\_SAM\_6AXIS (\emph{i-Octree}) shows surprising performance,  except for a  slight drift along the z-axis,  even without enabling loop closure.
LIO\_SAM\_6AXIS (static k-d tree) can only use the very recent key frames to estimate the poses,  whereas LIO\_SAM\_6AXIS  (\emph{i-Octree}) can take advantage of point information from the start, which contributes to its better performance.

\section{Conclusion}\label{sec:conclusion}
In this paper, we propose a novel dynamic octree data structure, \emph{i-Octree}, which supports incrementally point insertion with on-tree down-sampling, box-wise delete, and fast NNS.
Besides, a large amount of experiments on both randomized datasets and open datasets show that our \emph{i-Octree} can achieve the best overall performance among the state-of-the-art tree data structures.

\addtolength{\textheight}{-6cm}   % This command serves to balance the column lengths

\bibliography{ral-kdtree}

\end{document}